\documentclass[runningheads]{llncs}

\usepackage[T1]{fontenc}
\usepackage{graphicx}
\usepackage{amsmath}
\usepackage{amssymb}
\usepackage{booktabs}
\usepackage{bm}
\usepackage{wrapfig}
\usepackage{hyperref}

\begin{document}

\title{LCM-Promo: An LLM-Based Uplift Model for Promotion Allocation\thanks{This preprint has not undergone peer review or any post-submission improvements or corrections. The Version of Record of this contribution is published in \emph{Machine Learning and Knowledge Discovery in Databases. Applied Data Science Track, Demo Track and Industrial Track. ECML PKDD 2026}, and is available online at \protect\url{https://doi.org/10.1007/978-3-032-37685-5_53}.}}

\author{Nesta Midavaine\inst{1} \and
Floris Fok\inst{1} \and
Paulo Mendes\inst{2} \and
Joao Silva\inst{2} \and
\"Onder Aka{\c{c}}{\i}k\inst{1} \and
Dimas Lima\inst{2} \and
Z\"ulk\"uf Gen{\c{c}}\inst{1}}
\toctitle{LCM-Promo: An LLM-Based Uplift Model for Promotion Allocation}
\tocauthor{Nesta Midavaine, Floris Fok, Paulo Mendes, Joao Silva, \"Onder Aka{\c{c}}{\i}k, Dimas Lima, Z\"ulk\"uf Gen{\c{c}}}
\authorrunning{N. Midavaine et al.}

\institute{Prosus, Amsterdam, The Netherlands\\
\email{\{nesta.midavaine,floris.fok,onder.akacik,zulkuf.genc\}@prosus.com} \and
iFood, S\~ao Paulo, Brazil\\
\email{\{paulo.mendes,guedes.joao,dimas.lima\}@ifood.com.br}}

\maketitle

\begin{abstract}
We present LCM-Promo, an LLM-based uplift model for promotion allocation. LCM-Promo is a task-specific Large Commerce Model (LCM): a finetuned LLM trained on transactional commerce data to predict user intent and promotion response. It learns monotonic promotion-response curves from production logs by interpreting single-token outputs as conversion-threshold probabilities, while conditioning on session context such as day and delivery shift. In a production A/B test on churn reactivation, LCM-Promo increased conversion by 1\% and reduced promotion cost by 6\% relative to the Segment-EMA production baseline.
\end{abstract}

\section{Introduction}
\label{sec:intro}

LCM-Promo is one application of a broader effort at Prosus to build Large Commerce Models (LCMs): domain-specialized foundation models trained on user, merchant, and product interactions across e-commerce platforms. Where general-purpose LLMs reason about language, LCMs reason about commerce: user intent, merchant dynamics, market context, and responses to pricing and promotion. The LCM stack supports downstream tasks through direct serving and task-specific specialization. In this work we focus on promotion allocation.

Promotion targeting can be framed as uplift modelling: estimating how much more likely a user is to convert when shown a specific promotion than when shown no promotion, conditional on user and session features~\cite{zhang_unified_2021}. We use a finetuned LLM as an uplift model for promotion targeting in e-commerce. We deploy LCM-Promo at iFood, where promotions are ephemeral and context-dependent, varying by day, delivery shift, product category, and discount mechanic. This makes a prompt-based model a natural fit: user representations, session context, and discount options can be serialized directly into the input.

LCM-Promo predicts a single token over discount identifiers, which we interpret at decoding time as a distribution over user conversion thresholds. This yields a monotonic promotion-response curve by construction, usable both for budget-constrained allocation and for offline uplift evaluation. In a live A/B test on churn reactivation, LCM-Promo increased conversion by 1\% while reducing promotion cost by 6\% relative to the Segment-EMA production baseline, and was subsequently rolled out to all churn users. Offline, it is competitive with an uplift random forest trained on randomized control data.

\section{Method}
\label{sec:method}

We formulate promotion response estimation as single-token prediction. Each prompt contains a system prompt, session context, an LCM-generated user representation, and discount options mapped to alphabetic identifiers. The LCM representation model converts raw historical logs into structured behavioural summaries.

LCM-Promo uses a 4B parameter dense model as the base model architecture and predicts a single discount option through its identifier. Let $\mathcal{D}=\{d_0,\ldots,d_K\}$ be ordered by increasing discount depth, let $a_k$ be the identifier assigned to $d_k$, and let $a_\emptyset$ denote no purchase. The model is trained with cross-entropy on the observed conversion-discount label, using $a_\emptyset$ for non-converters. At inference time, we restrict the next-token distribution to these valid identifiers and renormalize:
\begin{equation}
    q_i(x)=
    \frac{p_\theta(a_i \mid x)}
    {\sum_{j=0}^{K}p_\theta(a_j \mid x)+p_\theta(a_\emptyset \mid x)},
    \qquad i\in\{0,\ldots,K,\emptyset\}.
    \label{eq:discount_pmf}
\end{equation}
We interpret this as a probability mass function (PMF) over a latent conversion threshold $T$, where $T=d_k$ means that $d_k$ is the minimum discount at which the user would convert, and $T=\infty$ corresponds to no purchase under any available discount. In our churn-reactivation setting, users may see multiple offers across one month but typically convert at most once, so the observed conversion discount is used as a proxy for this threshold. The predicted conversion probability for offer $d_k$ is the cumulative response curve,
\begin{equation}
    r_k(x)=P(T\leq d_k\mid x)=\sum_{j=0}^{k}q_j(x),
    \label{eq:response}
\end{equation}
which is monotone non-decreasing by construction. The no-purchase mass allows $r_K(x)<1$. Model-implied uplift relative to no discount follows directly.

We also evaluate a \textit{waterfall augmentation} that encodes the threshold ordering in the labels. If a user converts at $d_k$, all discounts $d_j\geq d_k$ are labelled positive; non-converters are positive only for $a_\emptyset$. These labels are smoothed into a soft target and trained with cross-entropy~\cite{bruch_analysis_2019}.

\section{Experiments}
\label{sec:experiments}

\subsection{Online Evaluation}
\label{subsec:online}

We compare against Segment-EMA, iFood's prior production policy, which forecasts orders per segment-promotion cell using an exponentially weighted moving average. Because each new discount, segment, or context requires exploration within the cell, Segment-EMA limits segmentation granularity. We conducted a one-month production A/B test on churn reactivation users across Brazil. Users were randomly assigned to a 90\% control group running Segment-EMA and a 10\% treatment group running LCM-Promo. Both arms used the same global budget constraint and the same linear-optimization allocator~\cite{albert2022commerce}. Only the response model differed.

\begin{wraptable}{r}{0.42\textwidth}
\centering
\caption{Online A/B test. Lifts relative to Segment-EMA baseline.}
\label{tab:online}
\begin{tabular}{lr}
\toprule
Metric & Relative lift \\
\midrule
Conversion & $+1\%$ \\
Promotion cost & $-6\%$ \\
\bottomrule
\end{tabular}
\end{wraptable}

LCM-Promo runs inference twice daily, for lunch and dinner. Users are grouped into city, session, and lifecycle clusters, and mean response curves are estimated by random sampling with bootstrap-determined sample sizes. Productionalization was made feasible by this sampling. Prompts use precomputed LCM-generated user representations and are submitted to a self-hosted vLLM deployment of the 4B parameter model via BESH~\cite{fok2025besh}, our open-source batch endpoint for self-hosted GPUs, running 1 hour daily on 8 H200s.

Because delivery shift and other session contexts are prompt inputs rather than separate Segment-EMA forecasting cells, LCM-Promo expanded allocation from about 50 segments to over 1{,}000 clusters and from 10 wallet configurations to over one million promotion combinations. Table~\ref{tab:online} shows a 1\% relative conversion lift and 6\% lower total promotion cost. The system was subsequently rolled out to all churn users.

\subsection{Offline Evaluation}
\label{subsec:offline}

Before deployment, we evaluated LCM-Promo on randomized controlled trial (RCT) data with a fixed promotion. We compare against a Uplift Random Forest implemented with \texttt{causalml}~\cite{soltys2015ensemble,chen2020causalml}, a strong baseline trained on previous months of RCT data. LCM-Promo is trained on historical production logs and is only evaluated on the RCT. This fixed-promotion benchmark is narrower than the production setting, which spans multiple promotion depths, session contexts, and wallet configurations. We therefore treat it as validation for online testing rather than evidence of offline superiority.

\begin{table}[t]
\centering
\caption{Offline uplift metrics on churn users.}
\label{tab:offline}
\begin{tabular}{lccc}
\toprule
Model & AUUC & Qini & Uplift@30 \\
\midrule
Uplift Random Forest          & $0.0578 \pm 0.0064$ & $0.0466 \pm 0.0053$ & $\bm{0.1708 \pm 0.0064}$ \\
LCM-Promo (post-processing)   & $\bm{0.0608 \pm 0.0032}$ & $\bm{0.0490 \pm 0.0028}$ & $0.1626 \pm 0.0035$ \\
LCM-Promo (waterfall)         & $0.0278 \pm 0.0036$ & $0.0226 \pm 0.0029$ & $0.1208 \pm 0.0026$ \\
\bottomrule
\end{tabular}
\end{table}

Table~\ref{tab:offline} shows that LCM-Promo is comparable on AUUC and Qini and below the random forest on Uplift@30. The waterfall variant performs worse, likely because its multi-label soft targets conflict with the LLM's next-token pretraining objective.

\section{Conclusion}
\label{sec:conclusion}

We presented LCM-Promo, a task-specific Large Commerce Model for promotion allocation. It learns promotion-response curves from production logs and converts them into monotonic uplift estimates through threshold-based PMF decoding. In a production A/B test on iFood churn reactivation, LCM-Promo increased conversion by 1\% and reduced promotion cost by 6\% relative to Segment-EMA, while expanding the allocation space by orders of magnitude. The system was subsequently rolled out to all churn users.

These results are limited to one lifecycle stage and one platform. Extending the approach to other lifecycle stages and commerce platforms is ongoing work. More broadly, LCM-Promo shows how Large Commerce Models can be specialized into compact production models for commerce decisions such as targeting, and merchant operations.

\begin{credits}
\subsubsection{\discintname}
The authors have no competing interests to declare.
\end{credits}


\begin{thebibliography}{1}
\providecommand{\url}[1]{\texttt{#1}}
\providecommand{\urlprefix}{URL }
\providecommand{\doi}[1]{https://doi.org/#1}

\bibitem{albert2022commerce}
Albert, J., Goldenberg, D.: E-commerce promotions personalization via online
  multiple-choice knapsack with uplift modeling. In: Proceedings of the 31st
  ACM International Conference on Information \& Knowledge Management. pp.
  2863--2872 (2022)

\bibitem{bruch_analysis_2019}
Bruch, S., Wang, X., Bendersky, M., Najork, M.: An {Analysis} of the {Softmax}
  {Cross} {Entropy} {Loss} for {Learning}-to-{Rank} with {Binary} {Relevance}.
  In: Proceedings of the 2019 {ACM} {SIGIR} {International} {Conference} on
  {Theory} of {Information} {Retrieval}. pp. 75--78. {ICTIR} '19, Association
  for Computing Machinery, New York, NY, USA (Sep 2019).
  \doi{10.1145/3341981.3344221}

\bibitem{chen2020causalml}
Chen, H., Harinen, T., Lee, J.Y., Yung, M., Zhao, Z.: Causalml: Python package
  for causal machine learning. arXiv preprint arXiv:2002.11631  (2020)

\bibitem{fok2025besh}
Fok, F., Aguiar, M., Borges, G., van~der Boor, P., Duncan, M.: {BESH}: An open
  source batch endpoint for unlocking 4x utilisation of your self hosted {GPU}
  cluster. Prosus AI Tech Blog (Oct 2025)

\bibitem{soltys2015ensemble}
So{\l}tys, M., Jaroszewicz, S., Rzepakowski, P.: Ensemble methods for uplift
  modeling. Data mining and knowledge discovery  \textbf{29},  1531--1559
  (2015)

\bibitem{zhang_unified_2021}
Zhang, W., Li, J., Liu, L.: A {Unified} {Survey} of {Treatment} {Effect}
  {Heterogeneity} {Modelling} and {Uplift} {Modelling}. ACM Comput. Surv.
  \textbf{54}(8),  162:1--162:36 (Oct 2021). \doi{10.1145/3466818}

\end{thebibliography}
\end{document}